\documentclass[11pt,a4paper]{article}
\usepackage[hyperref]{acl2021}
\usepackage{times}
\usepackage{latexsym}

\usepackage{microtype}
\usepackage{algorithm}
\usepackage{algpseudocode}
\usepackage{amsmath}
\usepackage{booktabs}
\usepackage{multirow}

\usepackage{subcaption}
\usepackage{pgfplots}
\pgfplotsset{compat=1.16}
\usepackage{tikz}
\usepackage{bm}
\usepackage{flushend}

\usetikzlibrary{patterns}

\aclfinalcopy 
\def\aclpaperid{4076} 

\DeclareMathOperator*{\argmax}{arg\,max}

\title{Improving Cross-Lingual Reading Comprehension with Self-Training}

\author{Wei-Cheng Huang$^*$ \\
  National Taiwan University \\
  \texttt{b05902015@ntu.edu.tw} \\\And
  Chien-yu Huang\thanks{$^*$ These authors contributed equally.} \\
  National Taiwan University \\
  \texttt{r08921062@ntu.edu.tw} \\\And
  Hung-yi Lee \\
  National Taiwan University \\
  \texttt{hungyilee@ntu.edu.tw}}

\date{}

\begin{document}
\maketitle
\begin{abstract}
Substantial improvements have been made in machine reading comprehension, where the machine answers questions based on a given context.
Current state-of-the-art models even surpass human performance on several benchmarks.
However, their abilities in the cross-lingual scenario are still to be explored.
Previous works have revealed the abilities of pre-trained multilingual models for zero-shot cross-lingual reading comprehension.
In this paper, we further utilized unlabeled data to improve the performance.
The model is first supervised-trained on source language corpus, and then self-trained with unlabeled target language data.
The experiment results showed improvements for all languages, and we also analyzed how self-training benefits cross-lingual reading comprehension in qualitative aspects.

\end{abstract}

\section{Introduction}
Machine reading comprehension has gained much attention because of its practical applications, and substantial improvements have been made recently.
Pre-trained models such as BERT~\cite{devlin-etal-2019-bert} and GPT family~\cite{radfordlanguage, brown2020language}  achieved state-of-the-art performance, or even surpassed the humans, in several benchmarks, including SQuAD~\cite{rajpurkar2016squad, rajpurkar2018know}, HotpotQA~\cite{yang2018hotpotqa}, and CoQA~\cite{reddy2019coqa}.

However, previous works mostly focused on monolingual reading comprehension, and large-scale data is usually available only in English.
Several multilingual datasets were then created for development~\cite{cui2019cross, liu2019xqa, lewis2019mlqa}.
Also, techniques such as transfer learning or domain adaptation thus become important for cross-lingual reading comprehension.
\citet{hsu2019zero} showed the ability of multilingual BERT (m-BERT) for zero-shot cross-lingual reading comprehension.
They first trained an m-BERT on source language reading comprehension corpus, and the model is then able to be applied to those in another target language without further fine-tuning.

On the other hand, self-training has been attractive in knowledge transfer.
In self-training, a model is first trained on a labeled dataset.
Then, the trained model gives pseudo-labels to unlabeled data.
A subset of these pseudo-labeled data is selected into the labeled dataset for the next training epoch.
This procedure is iterated several times.

This is the first paper using self-training to improve cross-lingual reading comprehension.
In real applications, we may still have access to sufficient target language data, but without well-annotated answers.
We thus propose to utilize these data by self-training.
Model is first trained on a source language corpus with annotation and then self-trained on target language data where only contexts and questions are available.
Pseudo-labels are dynamically obtained depending on the model's confidence during self-training.
Experiment results showed improvements among all corpora in four different languages, compared to previous zero-shot performance.
We also analyzed how self-training works for cross-lingual reading comprehension.

\section{Related Works}
\subsection{Cross-lingual transfer learning}
While there are hundreds of languages in the world, large-scale datasets that are crucial for machine learning, are usually only available in English.
Transfer learning for cross-lingual understanding thus becomes an important issue.
\citet{kim2017cross} utilized adversarial training to obtain language-general information without ancillary resources and achieved better performance on POS tagging.
\citet{schuster2019cross} used a machine translation encoder to obtain cross-lingual contextualized representations and it outperformed pre-trained multilingual ELMo~\cite{peters2018deep} on text classification and sequence labeling.

With the development of pre-trained models such as BERT~\cite{devlin-etal-2019-bert}, zero-shot cross-lingual transfer became more feasible.
Such cross-lingual transfer has been studied in several tasks like text classification~\cite{eriguchi2018zero}, dependency parsing~\cite{wang2019cross}, text summarization~\cite{duan2019zero}, and reading comprehension~\cite{hsu2019zero}.

\subsection{Self-training}

Self-training has succeeded in several domains.
\citet{zoph2020rethinking} showed the generality and flexibility of self-training and achieved better performance compared to pre-trained models in computer vision.
Such techniques have also been applied for speech recognition~\cite{kahn2020self, chen2020aipnet} and speech translation~\cite{pino2020self}.

In NLP, self-training has also been studied and applied in several aspects.
\citet{du2020self} showed that self-training is complementary to pre-training in standard text classification benchmarks.
\citet{dong2019robust} used self-training on cross-lingual document and sentiment classification.
However, the effectiveness of self-training on cross-lingual reading comprehension remained unknown, since it requires more reasoning and understanding compared to simple text classification, which led to the idea of self-training on cross-lingual reading comprehension as proposed here.

\section{Proposed Approach}

A well-labeled reading comprehension dataset $D$ is composed of at least three components: contexts $C = \{c_1, c_2, \ldots, c_n\}$, questions $Q = \{q_{11}, q_{12}, \ldots, q_{nm}\}$, and corresponding answers $A = \{a_{11}, a_{12}, \ldots, a_{nm}\}$, where $q_{ij}$ denotes the $j$-th question of $i$-th context and $a_{ij}$ is the corresponding answer.
In this paper, we assume that we have access to a labeled dataset $D_S = \{C_S, Q_S, A_S\}$ in source language $S$ and a unlabeled dataset $D_T = \{C_T, Q_T\}$ in target language $T$.
Given a pre-trained multilingual model, we first fine-tune it on the labeled dataset $D_S$ (fine-tuning stage), and then use $D_T$ for self-training (self-training stage).

\subsection{Self-training process}
\label{sec:self-training-process}
The complete two-stage process works as specified in Algorithm~\ref{alg:self-training}.
First, in fine-tuning stage, the pre-trained multilingual model $M_P$ is fine-tuned on the labeled source language dataset $D_S$ in a supervised manner.
The model, denoted as $M_0$, is then used as the starting point in the next stage.
In the self-training stage, the training and labeling process iterate several times.
For each iteration $i$, we first generate pseudo-labels for $D_T$ with model $M_i$ (Sec.~\ref{sec:labeling-process}), and train $M_0$ on $D_T$ with these labels.

\begin{algorithm}[htb]
    \caption{Cross-lingual self-training process.}
    \small
    \begin{algorithmic}[1]
        \Require
        \Statex $M_P$: Pre-trained multilingual model.
        \Statex $D_S$: Labeled dataset in the source language.
        \Statex $D_T$: Unlabeled dataset in the target language.
        \Statex $N$: Number of self-training iterations.
        \Statex $\theta$: Threshold for filtering pseudo-labels.
        \State $M_0 \leftarrow \text{train}(M_P, D_S)$
        \State $D_0 \leftarrow \text{label}(M_0, D_T, \theta)$
        \For{each $i \in [1, N]$}
            \State $M_i \leftarrow \text{train}(M_0, D_{i - 1})$
            \State $D_i \leftarrow \text{label}(M_i, D_T, \theta)$
        \EndFor
        \State \Return $M_N$
    \end{algorithmic}
    \label{alg:self-training}
\end{algorithm}

\subsection{Labeling process}
\label{sec:labeling-process}
To obtain the pseudo-labels, we run the unlabeled set through the model.
Algorithm~\ref{alg:pseudo-label} explains the labeling process.
For each unlabeled example $(c_i, q_{ij})$, the model gives $p$ answers $A_{ij} = \{a_{ij1}, a_{ij2}, \ldots, a_{ijp}\}$ with corresponding confidences $T_{ij} = \{\tau_{ij1}, \tau_{ij2}, \ldots, \tau_{ijp}\}$.
We then filter data by a threshold $\theta$.
Only examples whose greatest confidence exceeds the threshold are preserved.
For each question, the answer with the greatest confidence is used as the pseudo-label.

\begin{algorithm}[htb]
    \caption{Pseudo-label generation.}
    \small
    \begin{algorithmic}[1]
        \Require
        \Statex $M$: Model used in self-training process.
        \Statex $D_T$: Unlabeled dataset in the target language.
        \Statex $\theta$: Confidence threshold.
        \State $D^{'} \leftarrow \emptyset$
        \For{$c_i \in C_T$}
            \For{$q_{ij} \in Q_T$}
                \State $A_{ij}, T_{ij} \leftarrow M(c_i, q_{ij})$
                \If{$\max_{\tau}T_{ij} \geq \theta$}
                \State $k \leftarrow \argmax T_{ij}$
                \State $D^{'} \leftarrow D^{'} \cup \{(c_i, q_{ij}, a_{ijk})\}$
                \EndIf
            \EndFor
        \EndFor
        \State \Return $D^{'}$
    \end{algorithmic}
    \label{alg:pseudo-label}
\end{algorithm}

\section{Experimental Settings}
\subsection{Model}
We utilized publicly-available pre-trained m-BERT~\cite{wolf2020transformers} in our experiment.
The m-BERT is a transformer-based encoder pre-trained on multilingual Wikipedia and can perform well on several cross-lingual tasks.

\subsection{Datasets}
We used 4 datasets in different languages in the experiments: SQuAD v1.1~\cite{rajpurkar2016squad} in English, FQuAD~\cite{d2020fquad} in French, KorQuAD v1.0~\cite{lim2019korquad1} in Korean, and DRCD~\cite{shao2018drcd} in traditional Chinese.
These datasets were collected from Wikipedia in different languages, and all of the questions are answerable.
However, the test set of FQuAD is not available, and we thus evaluated our approaches on development sets for all languages.

\subsection{Pseudo-labeling}
In extraction-based reading comprehension, the model answers the question by yielding two probability distributions over the context tokens.
One is for the start position, and the other is for the end position.
The answer is then extracted by selecting the span between the start and end positions.

Here we consider the probabilities as a measure of confidence.
More specifically, we took the distributions of start and end tokens as input, then filtered out impossible candidates (for example, candidates with end token before start token), and performed beam search with the summation of start/end distributions as confidence.
The threshold $\theta$ was set to 0.7, which means only the examples whose most confident answer with a score greater than 0.7 were used in the self-training stage.

\begin{table*}[h]
    \centering
    \resizebox{\linewidth}{!}{
    \begin{tabular}{llcccc|llcccc}
        \toprule
        \multicolumn{2}{c}{\textbf{Dataset}} & \multicolumn{2}{c}{\textbf{EM}} & \multicolumn{2}{c}{\textbf{F1}} &
        \multicolumn{2}{c}{\textbf{Dataset}} & \multicolumn{2}{c}{\textbf{EM}} & \multicolumn{2}{c}{\textbf{F1}} \\
        \cmidrule(lr){1-2} \cmidrule(lr){3-4} \cmidrule(lr){5-6}
        \cmidrule(lr){7-8} \cmidrule(lr){9-10} \cmidrule(lr){11-12}
        Source & Target & \multirow{1}{*}{Zero-shot} & \multirow{1}{*}{Self-trained} & \multirow{1}{*}{Zero-shot} & \multirow{1}{*}{Self-trained} &
        Source & Target & \multirow{1}{*}{Zero-shot} & \multirow{1}{*}{Self-trained} & \multirow{1}{*}{Zero-shot} & \multirow{1}{*}{Self-trained} \\
        \midrule
        \multirow{3}{*}{SQuAD} & FQuAD & 53.39 & 57.59 & 72.78 & 75.75 & \multirow{3}{*}{DRCD} & SQuAD & 56.33 & 66.26 & 69.22 & 77.49 \\
        {} & DRCD & 67.00 & 71.25 & 82.94 & 86.65 & {} & FQuAD & 35.10 & 42.57 & 54.83 & 63.52 \\
        {} & KorQuAD & 53.22 & 61.07 & 75.63 & 83.99 & {} & KorQuAD & 56.62 & 61.71 & 79.77 & 85.70 \\
        \midrule
        \multirow{3}{*}{FQuAD} & SQuAD & 68.56 & 72.41 & 79.84 & 82.67 & \multirow{3}{*}{KorQuAD} & SQuAD & 61.69 & 67.01 & 71.74 & 76.80 \\
        {} & DRCD & 49.74 & 65.38 & 72.90 & 81.33 & {} & FQuAD & 39.46 & 42.35 & 56.24 & 59.02 \\
        {} & KorQuAD & 55.68 & 58.76 & 73.62 & 82.17 & {} & DRCD & 76.14 & 77.89 & 84.11 & 85.77 \\
        \bottomrule
    \end{tabular}
    }
    \caption{
        Comparison of the performance (EM and F1) between zero-shot and self-trained m-BERTs on different datasets.
        The pre-trained m-BERT was first fine-tuned on the source language, and then self-trained and evaluated on the target language.
    }
    \label{tab:all-pairs}
\end{table*}

\section{Results}

\subsection{Quantitative results}
Following the convention, we used exact match (EM) and F1 scores to evaluate the performance.
Since we used 4 datasets in the experiments, there are therefore 12 \textit{source-target} pairs to be evaluated.
The notation \textit{source-target} denotes that the model was fine-tuned on the source dataset, and then self-trained and evaluated on the target dataset.
Table~\ref{tab:all-pairs} presents the results on all 12 pairs across the four datasets.
The m-BERT was first trained on the source language dataset, self-trained on the training set of the target language, and evaluated on the development set of the target language.
Zero-shot means the m-BERT is only trained on the source language without self-training.
We see that both EM and F1 were enhanced across all pairs with the help of self-training, compared to the original zero-shot ones (+1.75 to +15.64 on EM and +1.66 to +8.69 on F1).
The consistent improvements show that our approach is general but not limited to specific languages.
Table~\ref{tab:squad-other-f1-iterations} lists the F1 scores in each iteration when m-BERT was first fine-tuned on SQuAD and self-trained on other datasets.
In each iteration, almost all models were made better, and the first iteration benefited the most, as there are obvious gaps between zero-shot and iteration 1 for all models on both EM and F1 scores.

\begin{table}[htb]
    \centering
    \small
    \begin{tabular}{ccccc}
        \toprule
        \textbf{Target} & Zero-shot & Iter. 1 & Iter. 2 & Iter. 3 \\
        \midrule
        FQuAD & 72.78 & 74.19 & 74.96 & 75.75 \\
        DRCD & 82.94 & 85.55 & 85.59 & 86.64 \\
        KorQuAD & 75.63 & 82.00 & 83.27 & 83.99 \\
        \bottomrule
    \end{tabular}
    \caption{Comparison of F1 scores of each iteration when m-BERT was first fine-tuned on SQuAD and self-trained on different datasets.}
    \label{tab:squad-other-f1-iterations}
\end{table}

\subsection{Qualitative analysis}

We then investigated how our approach enhances the performance.
Specifically, we studied the cross-lingual transfer on SQuAD, DRCD, and FQuAD in terms of the question and answer types.
We classified questions and answers with the same procedure mentioned in their original papers, and the details are described in the appendix.
KorQuAD was excluded here because the question and answer types were not specified clearly in the original paper, and we did not found a publicly-available reliable tagging toolkit for Korean.

\begin{figure}[htb]
    \centering
    \includegraphics[width=.9\linewidth]{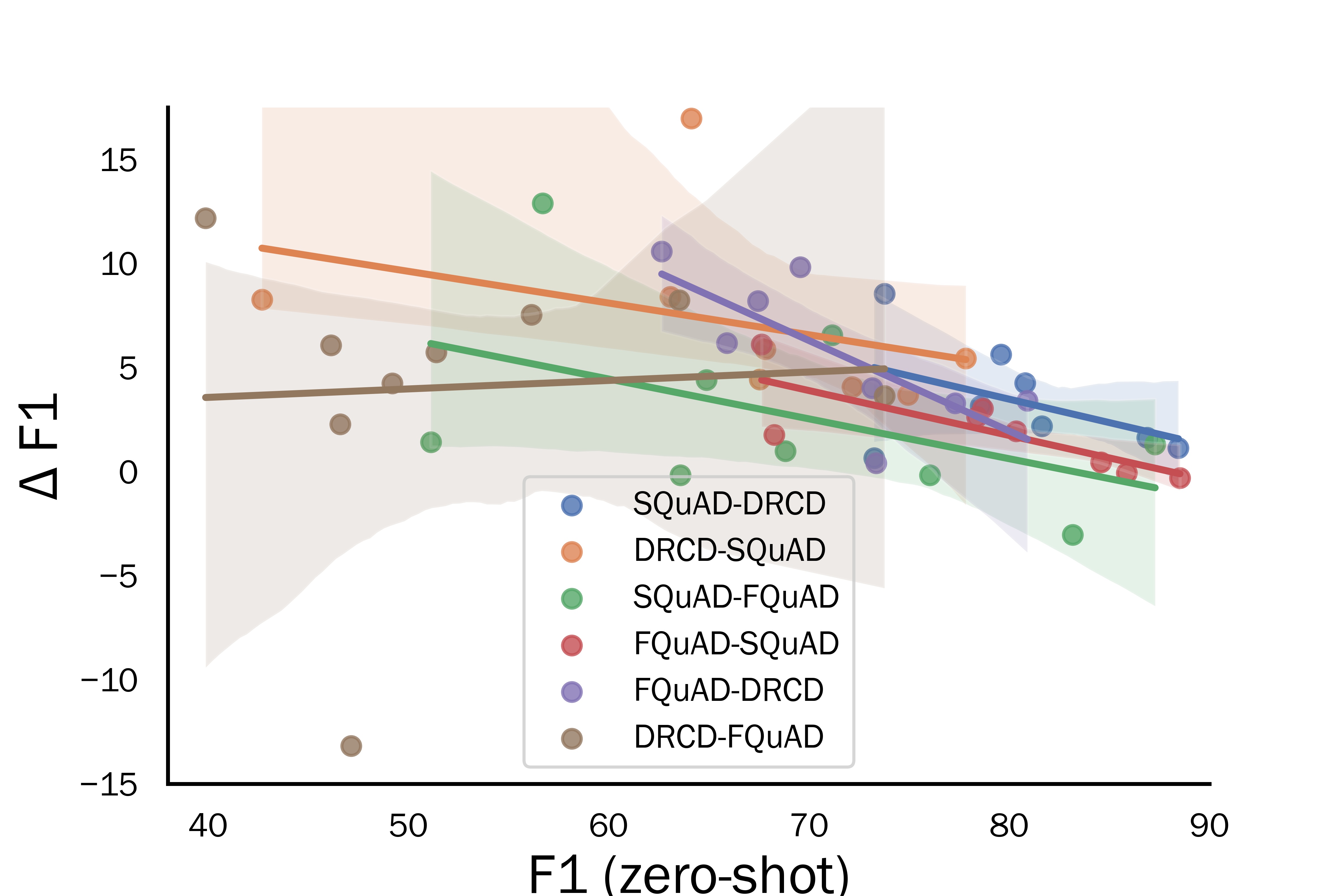}
    \caption{The zero-shot F1 scores with corresponding improvements in first iteration in self-training stage for different types of questions.}
    \label{fig:f1-diff-zero-shot}
\end{figure}

We first analyzed how the scores varied concerning its original zero-shot scores.
We wondered whether the larger amount of labels brings more benefits to their corresponding types of questions.
However, we did not see an obvious trend here, which indicates that the amount of pseudo-labels is not the key to the improvements for different question types (refer to the appendix for the figures).

We then looked into the relation between the zero-shot scores and corresponding improvements for different types of questions in the first iteration of self-training, as shown in Figure~\ref{fig:f1-diff-zero-shot}.
We see that self-training mostly improved the examples that were not answered well in the zero-shot inference (low F1 score on $x$-axis with good improvement on $y$-axis).
We also present the results for answer types but in the appendix due to the space limit.

\begin{table}[tb]
    \centering
    \small
        \begin{tabular}{l|cc}
        \toprule
         \textbf{Question Types} & DRCD-SQuAD & FQuAD-SQuAD \\
         \midrule
         Who & 5.45 & -0.3 \\
         What & 5.89 & 1.95 \\
         When & 16.98 & -0.05 \\
         Where & 4.44 & 3.04 \\
         Why & 8.27 & 1.77 \\
         Which & 4.07 & 0.46 \\
         How & 8.41 & 6.12 \\
         Other & 3.70 & 2.63 \\
         \bottomrule
    \end{tabular}
    \caption{The F1 improvements ($\Delta$F1) of each question type in the first iteration when self-trained on SQuAD.}
    \label{tab:other-squad-main}
\end{table}

\begin{table}[tb]
    \centering
    \small
    \begin{tabular}{l|cc}
        \toprule
        \textbf{Question Types} & SQuAD-DRCD & FQuAD-DRCD \\
        \midrule
        Who & 4.26 & 9.83 \\
        What & 3.16 & 8.21 \\
        When & 5.64 & 10.59 \\
        Where & 1.14 & 3.41 \\
        Why & 0.66 & 0.42 \\
        Which & 1.64 & 3.3 \\
        How & 8.55 & 6.19 \\
        Other & 2.19 & 4.02 \\
        \bottomrule
    \end{tabular}
    \caption{The F1 improvements ($\Delta$F1) of each question type in the first iteration when self-trained on DRCD.}
    \label{tab:other-drcd-main}
\end{table}

\begin{table}[tb]
    \centering
    \small
    \begin{tabular}{l|cc}
        \toprule
         \textbf{Question Types} & SQuAD-FQuAD & DRCD-FQuAD \\
         \midrule
         Who & 1.32 & 3.65 \\
         What (quoi) & -0.16 & 6.08 \\
         What (que) & -0.16 & 7.54 \\
         When & -3.03 & 4.25 \\
         Where & 12.90 & 2.28 \\
         Why & 1.42 & -13.17 \\
         How & 4.41 & 12.18 \\
         How many & 6.57 & 8.24 \\
         Other & 1.00 & 5.74 \\
         \bottomrule
    \end{tabular}
    \caption{The F1 improvements ($\Delta$F1) of each question type in the first iteration when self-trained on FQuAD.}
    \label{tab:other-fquad-main}
\end{table}

We further compared the improvements of each question and answer type in the first iteration.
Table~\ref{tab:other-squad-main},~\ref{tab:other-drcd-main} and~\ref{tab:other-fquad-main} show the results for question types, and those for answer types are in the appendix.
The question types vary in the tables due to different categories in their original papers.
We found that in the SQuAD-DRCD pair, "How" and "When" for questions and "Numeric" for answers improved the most in terms of F1 scores.
Similar results were obtained in the DRCD-SQuAD pair with "When" and "How" questions and "Date" and "Verb" answers.
In the FQuAD-SQuAD pair, "How" questions and "Adjective" and "Other Numeric" answers improved the most, but "Where" improved instead of "When".
This suggests our approach helps with the numeric questions and date answers the most.
Also, such improvements between DRCD and SQuAD were more than those between FQuAD and SQuAD.
We believe this is because their formats in Chinese much differs from those in English.

\section{Conclusion}
This paper presents the first self-training approach to improve cross-lingual machine reading comprehension.
The experiments were conducted on large-scale datasets in four different languages.
The results showed that our approach improved the performance significantly compared to the baseline with 1 -- 16 EM and 1 -- 8 F1 scores.
We also analyzed how self-training improves cross-lingual reading comprehension in several aspects and found that improvements are correlated to zero-shot performance but not the number of pseudo-labels.

\section*{Acknowledgments}
We are grateful to Taiwan Computing Cloud (\url{https://www.twcc.ai/}) for providing computation resources.

\bibliographystyle{acl_natbib}
\bibliography{anthology,acl2021}

\begin{thebibliography}{29}
\expandafter\ifx\csname natexlab\endcsname\relax\def\natexlab#1{#1}\fi

\bibitem[{Brown et~al.(2020)Brown, Mann, Ryder, Subbiah, Kaplan, Dhariwal,
  Neelakantan, Shyam, Sastry, Askell et~al.}]{brown2020language}
Tom~B Brown, Benjamin Mann, Nick Ryder, Melanie Subbiah, Jared Kaplan, Prafulla
  Dhariwal, Arvind Neelakantan, Pranav Shyam, Girish Sastry, Amanda Askell,
  et~al. 2020.
\newblock Language models are few-shot learners.
\newblock \emph{arXiv preprint arXiv:2005.14165}.

\bibitem[{Chen et~al.(2020)Chen, Yang, Yeh, Jain, and Seltzer}]{chen2020aipnet}
Yi-Chen Chen, Zhaojun Yang, Ching-Feng Yeh, Mahaveer Jain, and Michael~L
  Seltzer. 2020.
\newblock Aipnet: Generative adversarial pre-training of accent-invariant
  networks for end-to-end speech recognition.
\newblock In \emph{ICASSP 2020-2020 IEEE International Conference on Acoustics,
  Speech and Signal Processing (ICASSP)}, pages 6979--6983. IEEE.

\bibitem[{Cui et~al.(2019)Cui, Che, Liu, Qin, Wang, and Hu}]{cui2019cross}
Yiming Cui, Wanxiang Che, Ting Liu, Bing Qin, Shijin Wang, and Guoping Hu.
  2019.
\newblock Cross-lingual machine reading comprehension.
\newblock In \emph{Proceedings of the 2019 Conference on Empirical Methods in
  Natural Language Processing and the 9th International Joint Conference on
  Natural Language Processing (EMNLP-IJCNLP)}, pages 1586--1595.

\bibitem[{Devlin et~al.(2019)Devlin, Chang, Lee, and
  Toutanova}]{devlin-etal-2019-bert}
Jacob Devlin, Ming-Wei Chang, Kenton Lee, and Kristina Toutanova. 2019.
\newblock \href {https://doi.org/10.18653/v1/N19-1423} {{BERT}: Pre-training of
  deep bidirectional transformers for language understanding}.
\newblock In \emph{Proceedings of the 2019 Conference of the North {A}merican
  Chapter of the Association for Computational Linguistics: Human Language
  Technologies, Volume 1 (Long and Short Papers)}, pages 4171--4186,
  Minneapolis, Minnesota. Association for Computational Linguistics.

\bibitem[{d'Hoffschmidt et~al.(2020)d'Hoffschmidt, Vidal, Belblidia, and
  Brendl{\'e}}]{d2020fquad}
Martin d'Hoffschmidt, Maxime Vidal, Wacim Belblidia, and Tom Brendl{\'e}. 2020.
\newblock Fquad: French question answering dataset.
\newblock \emph{arXiv preprint arXiv:2002.06071}.

\bibitem[{Dong and de~Melo(2019)}]{dong2019robust}
Xin~Luna Dong and Gerard de~Melo. 2019.
\newblock A robust self-learning framework for cross-lingual text
  classification.
\newblock In \emph{Proceedings of the 2019 Conference on Empirical Methods in
  Natural Language Processing and the 9th International Joint Conference on
  Natural Language Processing (EMNLP-IJCNLP)}, pages 6307--6311.

\bibitem[{Du et~al.(2020)Du, Grave, Gunel, Chaudhary, Celebi, Auli, Stoyanov,
  and Conneau}]{du2020self}
Jingfei Du, Edouard Grave, Beliz Gunel, Vishrav Chaudhary, Onur Celebi, Michael
  Auli, Ves Stoyanov, and Alexis Conneau. 2020.
\newblock Self-training improves pre-training for natural language
  understanding.
\newblock \emph{arXiv preprint arXiv:2010.02194}.

\bibitem[{Duan et~al.(2019)Duan, Yin, Zhang, Chen, and Luo}]{duan2019zero}
Xiangyu Duan, Mingming Yin, Min Zhang, Boxing Chen, and Weihua Luo. 2019.
\newblock Zero-shot cross-lingual abstractive sentence summarization through
  teaching generation and attention.
\newblock In \emph{Proceedings of the 57th Annual Meeting of the Association
  for Computational Linguistics}, pages 3162--3172.

\bibitem[{Eriguchi et~al.(2018)Eriguchi, Johnson, Firat, Kazawa, and
  Macherey}]{eriguchi2018zero}
Akiko Eriguchi, Melvin Johnson, Orhan Firat, Hideto Kazawa, and Wolfgang
  Macherey. 2018.
\newblock Zero-shot cross-lingual classification using multilingual neural
  machine translation.
\newblock \emph{arXiv preprint arXiv:1809.04686}.

\bibitem[{Hsu et~al.(2019)Hsu, Liu, and Lee}]{hsu2019zero}
Tsung-Yuan Hsu, Chi-Liang Liu, and Hung-yi Lee. 2019.
\newblock Zero-shot reading comprehension by cross-lingual transfer learning
  with multi-lingual language representation model.
\newblock In \emph{Proceedings of the 2019 Conference on Empirical Methods in
  Natural Language Processing and the 9th International Joint Conference on
  Natural Language Processing (EMNLP-IJCNLP)}, pages 5935--5942.

\bibitem[{Kahn et~al.(2020)Kahn, Lee, and Hannun}]{kahn2020self}
Jacob Kahn, Ann Lee, and Awni Hannun. 2020.
\newblock Self-training for end-to-end speech recognition.
\newblock In \emph{ICASSP 2020-2020 IEEE International Conference on Acoustics,
  Speech and Signal Processing (ICASSP)}, pages 7084--7088. IEEE.

\bibitem[{Kim et~al.(2017)Kim, Kim, Sarikaya, and
  Fosler-Lussier}]{kim2017cross}
Joo-Kyung Kim, Young-Bum Kim, Ruhi Sarikaya, and Eric Fosler-Lussier. 2017.
\newblock Cross-lingual transfer learning for pos tagging without cross-lingual
  resources.
\newblock In \emph{Proceedings of the 2017 conference on empirical methods in
  natural language processing}, pages 2832--2838.

\bibitem[{Lewis et~al.(2019)Lewis, O{\u{g}}uz, Rinott, Riedel, and
  Schwenk}]{lewis2019mlqa}
Patrick Lewis, Barlas O{\u{g}}uz, Ruty Rinott, Sebastian Riedel, and Holger
  Schwenk. 2019.
\newblock Mlqa: Evaluating cross-lingual extractive question answering.
\newblock \emph{arXiv preprint arXiv:1910.07475}.

\bibitem[{Lim et~al.(2019)Lim, Kim, and Lee}]{lim2019korquad1}
Seungyoung Lim, Myungji Kim, and Jooyoul Lee. 2019.
\newblock Korquad1. 0: Korean qa dataset for machine reading comprehension.
\newblock \emph{arXiv preprint arXiv:1909.07005}.

\bibitem[{Liu et~al.(2019)Liu, Lin, Liu, and Sun}]{liu2019xqa}
Jiahua Liu, Yankai Lin, Zhiyuan Liu, and Maosong Sun. 2019.
\newblock Xqa: A cross-lingual open-domain question answering dataset.
\newblock In \emph{Proceedings of the 57th Annual Meeting of the Association
  for Computational Linguistics}, pages 2358--2368.

\bibitem[{Loshchilov and Hutter(2018)}]{loshchilov2018decoupled}
Ilya Loshchilov and Frank Hutter. 2018.
\newblock Decoupled weight decay regularization.
\newblock In \emph{International Conference on Learning Representations}.

\bibitem[{Manning et~al.(2014)Manning, Surdeanu, Bauer, Finkel, Bethard, and
  McClosky}]{manning2014stanford}
Christopher~D Manning, Mihai Surdeanu, John Bauer, Jenny~Rose Finkel, Steven
  Bethard, and David McClosky. 2014.
\newblock The stanford corenlp natural language processing toolkit.
\newblock In \emph{Proceedings of 52nd annual meeting of the association for
  computational linguistics: system demonstrations}, pages 55--60.

\bibitem[{Peters et~al.(2018)Peters, Neumann, Iyyer, Gardner, Clark, Lee, and
  Zettlemoyer}]{peters2018deep}
Matthew Peters, Mark Neumann, Mohit Iyyer, Matt Gardner, Christopher Clark,
  Kenton Lee, and Luke Zettlemoyer. 2018.
\newblock Deep contextualized word representations.
\newblock In \emph{Proceedings of the 2018 Conference of the North American
  Chapter of the Association for Computational Linguistics: Human Language
  Technologies, Volume 1 (Long Papers)}, pages 2227--2237.

\bibitem[{Pino et~al.(2020)Pino, Xu, Ma, Dousti, and Tang}]{pino2020self}
Juan Pino, Qiantong Xu, Xutai Ma, Mohammad~Javad Dousti, and Yun Tang. 2020.
\newblock Self-training for end-to-end speech translation.
\newblock \emph{Proc. Interspeech 2020}, pages 1476--1480.

\bibitem[{Radford et~al.()Radford, Wu, Child, Luan, Amodei, and
  Sutskever}]{radfordlanguage}
Alec Radford, Jeffrey Wu, Rewon Child, David Luan, Dario Amodei, and Ilya
  Sutskever.
\newblock Language models are unsupervised multitask learners.

\bibitem[{Rajpurkar et~al.(2018)Rajpurkar, Jia, and Liang}]{rajpurkar2018know}
Pranav Rajpurkar, Robin Jia, and Percy Liang. 2018.
\newblock Know what you don’t know: Unanswerable questions for squad.
\newblock In \emph{Proceedings of the 56th Annual Meeting of the Association
  for Computational Linguistics (Volume 2: Short Papers)}, pages 784--789.

\bibitem[{Rajpurkar et~al.(2016)Rajpurkar, Zhang, Lopyrev, and
  Liang}]{rajpurkar2016squad}
Pranav Rajpurkar, Jian Zhang, Konstantin Lopyrev, and Percy Liang. 2016.
\newblock Squad: 100,000+ questions for machine comprehension of text.
\newblock In \emph{Proceedings of the 2016 Conference on Empirical Methods in
  Natural Language Processing}, pages 2383--2392.

\bibitem[{Reddy et~al.(2019)Reddy, Chen, and Manning}]{reddy2019coqa}
Siva Reddy, Danqi Chen, and Christopher~D Manning. 2019.
\newblock Coqa: A conversational question answering challenge.
\newblock \emph{Transactions of the Association for Computational Linguistics},
  7:249--266.

\bibitem[{Schuster et~al.(2019)Schuster, Gupta, Shah, and
  Lewis}]{schuster2019cross}
Sebastian Schuster, Sonal Gupta, Rushin Shah, and Mike Lewis. 2019.
\newblock Cross-lingual transfer learning for multilingual task oriented
  dialog.
\newblock In \emph{Proceedings of the 2019 Conference of the North American
  Chapter of the Association for Computational Linguistics: Human Language
  Technologies, Volume 1 (Long and Short Papers)}, pages 3795--3805.

\bibitem[{Shao et~al.(2018)Shao, Liu, Lai, Tseng, and Tsai}]{shao2018drcd}
Chih~Chieh Shao, Trois Liu, Yuting Lai, Yiying Tseng, and Sam Tsai. 2018.
\newblock Drcd: a chinese machine reading comprehension dataset.
\newblock \emph{arXiv preprint arXiv:1806.00920}.

\bibitem[{Wang et~al.(2019)Wang, Che, Guo, Liu, and Liu}]{wang2019cross}
Yuxuan Wang, Wanxiang Che, Jiang Guo, Yijia Liu, and Ting Liu. 2019.
\newblock Cross-lingual bert transformation for zero-shot dependency parsing.
\newblock In \emph{Proceedings of the 2019 Conference on Empirical Methods in
  Natural Language Processing and the 9th International Joint Conference on
  Natural Language Processing (EMNLP-IJCNLP)}, pages 5725--5731.

\bibitem[{Wolf et~al.(2020)Wolf, Chaumond, Debut, Sanh, Delangue, Moi, Cistac,
  Funtowicz, Davison, Shleifer et~al.}]{wolf2020transformers}
Thomas Wolf, Julien Chaumond, Lysandre Debut, Victor Sanh, Clement Delangue,
  Anthony Moi, Pierric Cistac, Morgan Funtowicz, Joe Davison, Sam Shleifer,
  et~al. 2020.
\newblock Transformers: State-of-the-art natural language processing.
\newblock In \emph{Proceedings of the 2020 Conference on Empirical Methods in
  Natural Language Processing: System Demonstrations}, pages 38--45.

\bibitem[{Yang et~al.(2018)Yang, Qi, Zhang, Bengio, Cohen, Salakhutdinov, and
  Manning}]{yang2018hotpotqa}
Zhilin Yang, Peng Qi, Saizheng Zhang, Yoshua Bengio, William Cohen, Ruslan
  Salakhutdinov, and Christopher~D Manning. 2018.
\newblock Hotpotqa: A dataset for diverse, explainable multi-hop question
  answering.
\newblock In \emph{Proceedings of the 2018 Conference on Empirical Methods in
  Natural Language Processing}, pages 2369--2380.

\bibitem[{Zoph et~al.(2020)Zoph, Ghiasi, Lin, Cui, Liu, Cubuk, and
  Le}]{zoph2020rethinking}
Barret Zoph, Golnaz Ghiasi, Tsung-Yi Lin, Yin Cui, Hanxiao Liu, Ekin~Dogus
  Cubuk, and Quoc Le. 2020.
\newblock Rethinking pre-training and self-training.
\newblock \emph{Advances in Neural Information Processing Systems}, 33.

\end{thebibliography}

\newpage
\appendix

\section{Implementation Details}
For each experiment, we first fine-tuned m-BERT on the training set in the source language for 3 epochs.
Then, the training set in the target language was used as the unlabeled set for self-training for 3 iterations.
Then, in each iteration, the model was also trained for 3 epochs.
Last, the performance was evaluated on the development set in the target language.

For both fine-tuning and self-training stage, we trained the model using AdamW~\cite{loshchilov2018decoupled} with the linear scheduler without warm-up.
The batch size was 32, and the learning rate was $5 \times 10^{-5}$.

The maximum input context and question length of the model was set to 384 and 64 respectively, and the document stride was 128.
During inference, the maximum length of machine-generated predictions was 30, and we used a beam size of 20 when computing predictions.

\section{Classifying Questions and Answers}
\paragraph{SQuAD}
We classified the questions based on their word: What, When, Where, Which, Why, Who, How, and other.
As for the answers, following \citet{rajpurkar2016squad}, we leveraged constituency parses, POS tags, and NER tags with Stanford CoreNLP~\cite{manning2014stanford}.

\paragraph{DRCD}
For DRCD, we mostly followed \citet{shao2018drcd} to categorize the questions and answers.
However, we used our own set of keywords and measure words because the keywords and the measure words were not given in the original paper.

\paragraph{FQuAD}
We grouped the questions in the development set of FQuAD by their beginning interrogative words, the same as \citet{d2020fquad} did.
Nonetheless, since the regular expressions and rules to classify the answers were not specified in their paper, we followed a similar procedure in SQuAD to categorize them.

\section{Improvement Analysis}

\begin{figure}[H]
    \centering
    \includegraphics[width=\linewidth]{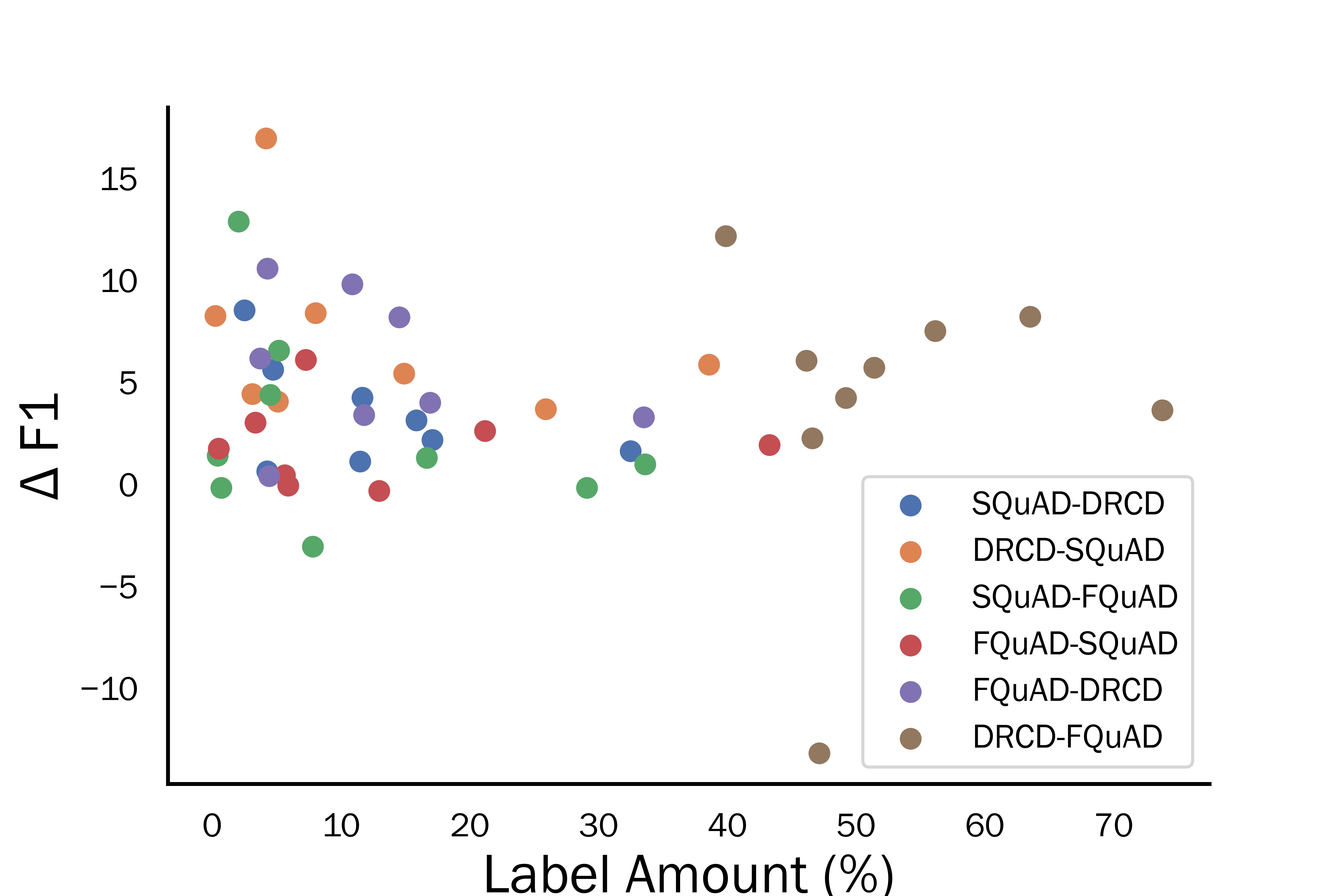}
    \caption{The relation between F1 improvements and the amount of pseudo-labels for different question types in the first iteration of self-training stage.}
    \label{fig:f1-diff-amount}
\end{figure}

\begin{figure}[H]
    \centering
    \includegraphics[width=\linewidth]{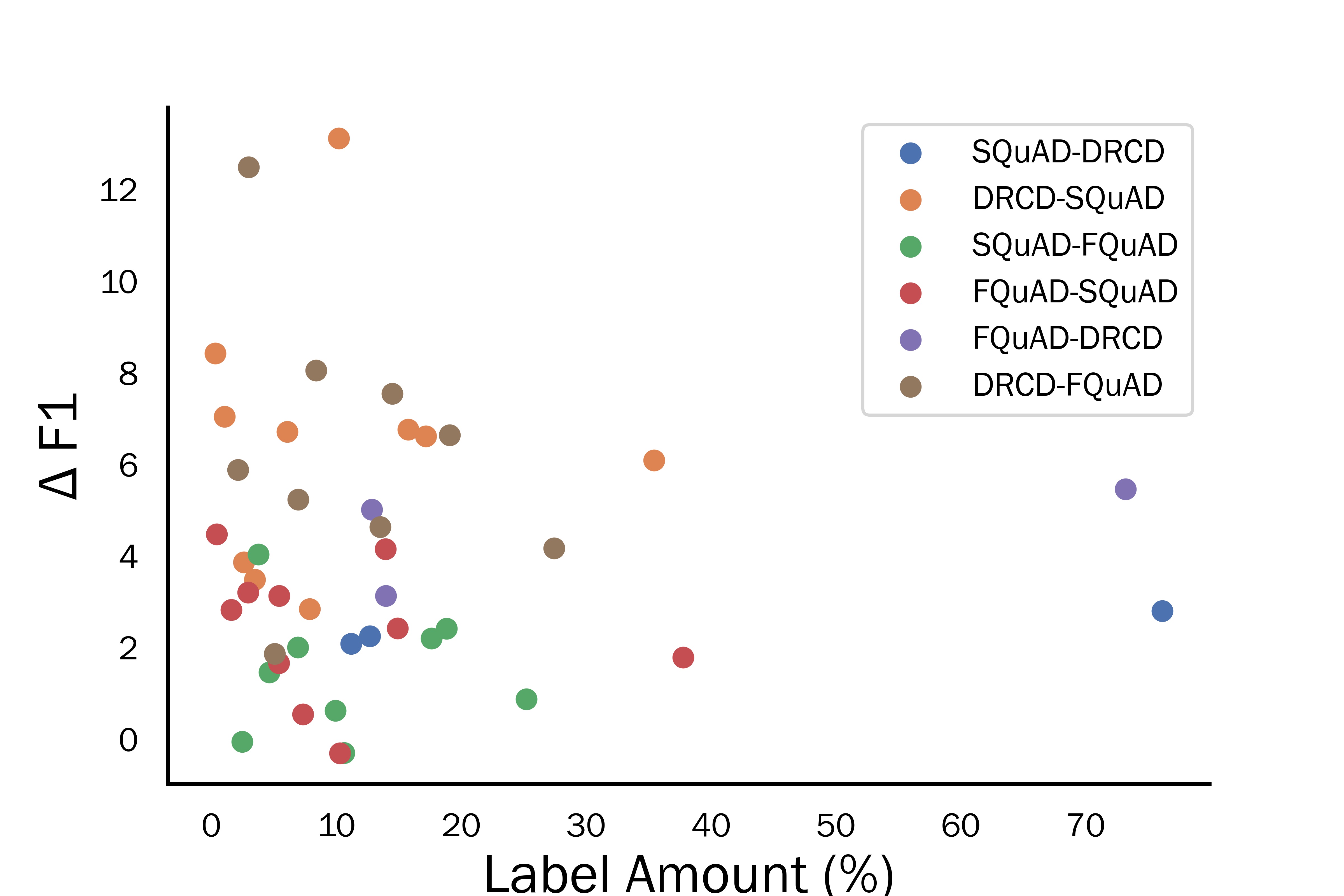}
    \caption{The relation between F1 improvements and the amount of pseudo-labels for different answer types in the first iteration of self-training stage.}
    \label{fig:ans-label-amount}
\end{figure}

\begin{figure}[H]
    \centering
    \includegraphics[width=\linewidth]{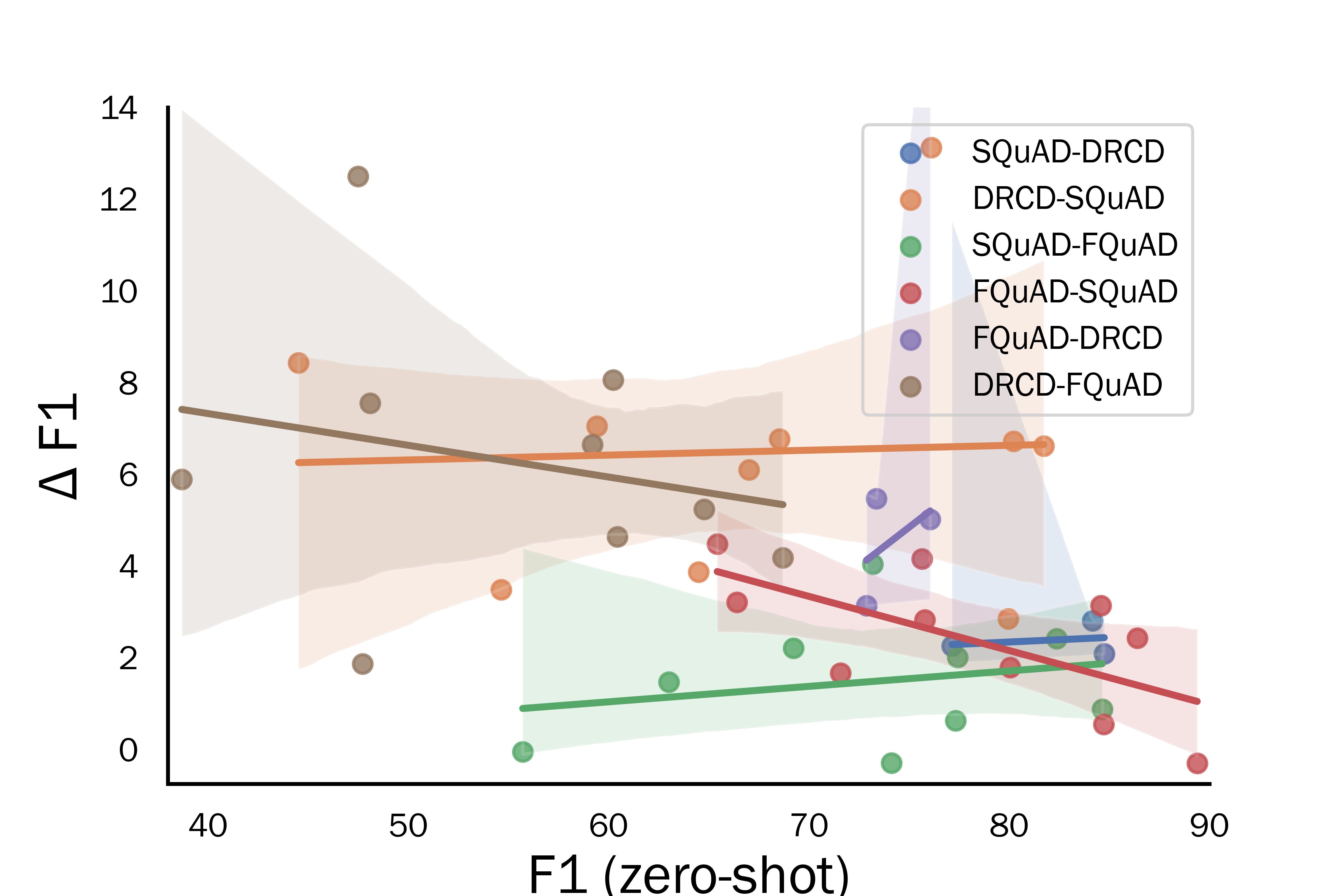}
    \caption{The change of F1 scores in the first iteration in self-training stage for different types of answers.}
    \label{fig:f1-diff-ans}
\end{figure}

\begin{table}[]
    \centering
    \resizebox{\linewidth}{!}{
    \footnotesize
        \begin{tabular}{l|cc}
        \toprule
         \textbf{Types} & DRCD-SQuAD & FQuAD-SQuAD \\
         \midrule
         \multicolumn{3}{l}{\textbf{Questions}} \\
         \midrule
         Who & 5.45 & -0.3 \\
         What & 5.89 & 1.95 \\
         When & 16.98 & -0.05 \\
         Where & 4.44 & 3.04 \\
         Why & 8.27 & 1.77 \\
         Which & 4.07 & 0.46 \\
         How & 8.41 & 6.12 \\
         Other & 3.70 & 2.63 \\
         \midrule
         \multicolumn{3}{l}{\textbf{Answers}} \\
         \midrule
         Date & 13.13 & -0.3 \\
         Other Numeric & 6.77 & 4.16 \\
         Person & 6.62 & 2.43 \\
         Location & 6.72 & 3.14 \\
         Other Entity & 2.85 & 0.55 \\
         Common Noun Phrase & 6.10 & 1.79 \\
         Verb Phrase & 8.43 & 4.48 \\
         Adjective Phrase & 3.87 & 3.21 \\
         Clause & 3.49 & 1.67 \\
         Other & 7.05 & 2.83 \\
         \bottomrule
    \end{tabular}
    }
    \caption{The F1 improvements ($\Delta$F1) of each question and answer type in the first iteration when self-trained on SQuAD.}
    \label{tab:other-squad-appendix}
\end{table}

\begin{table}[]
    \centering
    \footnotesize
    \begin{tabular}{l|cc}
        \toprule
        \textbf{Types} & SQuAD-DRCD & FQuAD-DRCD \\
        \midrule
        \multicolumn{3}{l}{\textbf{Questions}} \\
        \midrule
        Who & 4.26 & 9.83 \\
        What & 3.16 & 8.21 \\
        When & 5.64 & 10.59 \\
        Where & 1.14 & 3.41 \\
        Why & 0.66 & 0.42 \\
        Which & 1.64 & 3.3 \\
        How & 8.55 & 6.19 \\
        Other & 2.19 & 4.02 \\
        \midrule
        \multicolumn{3}{l}{\textbf{Answers}} \\
        \midrule
        Numeric & 2.09 & 5.02 \\
        Entity & 2.81 & 5.47 \\
        Description & 2.26 & 3.14 \\
        \bottomrule
    \end{tabular}
    \caption{The F1 improvements ($\Delta$F1) of each question and answer type in the first iteration when self-trained on DRCD.}
    \label{tab:other-drcd-appendix}
\end{table}

\begin{table}[]
    \centering
    \resizebox{\linewidth}{!}{
    \footnotesize
    \begin{tabular}{l|cc}
        \toprule
         \textbf{Types} & SQuAD-FQuAD & DRCD-FQuAD \\
         \midrule
         \multicolumn{3}{l}{\textbf{Questions}} \\
         \midrule
         Who & 1.32 & 3.65 \\
         What (quoi) & -0.16 & 6.08 \\
         What (que) & -0.16 & 7.54 \\
         When & -3.03 & 4.25 \\
         Where & 12.90 & 2.28 \\
         Why & 1.42 & -13.17 \\
         How & 4.41 & 12.18 \\
         How many & 6.57 & 8.24 \\
         Other & 1.00 & 5.74 \\
         \midrule
         \multicolumn{3}{l}{\textbf{Answers}} \\
         \midrule
         Date & 2.42 & 6.65 \\
         Other Numeric & -0.29 & 4.64 \\
         Person & 0.88 & 4.18 \\
         Location & 0.63 & 8.06 \\
         Other Proper Nouns & 2.01 & 5.24 \\
         Common Noun & 2.21 & 7.55 \\
         Verb & -0.05 & 5.89 \\
         Adjective & 4.04 & 12.5 \\
         Other & 1.47 & 1.87 \\
         \bottomrule
    \end{tabular}
    }
    \caption{The F1 improvements ($\Delta$F1) of each question and answer type in the first iteration when self-trained on FQuAD.}
    \label{tab:other-fquad-appendix}
\end{table}

Figure~\ref{fig:f1-diff-amount} and Figure~\ref{fig:ans-label-amount} respectively show the relations between F1 improvements and the number of pseudo-labels of different question or answer types in the first self-training iteration.
The numbers and content of types vary due to different categories in their original papers.
We can see that the number of pseudo-labels does not strongly related to the improvements on F1 score.
On the other hand, Figure~\ref{fig:f1-diff-ans} displays the improvements of F1 scores for different types of answers.
It shows a similar trend with those in question types, but is not as obvious.

Table~\ref{tab:other-squad-appendix},~\ref{tab:other-drcd-appendix} and~\ref{tab:other-fquad-appendix} show the improvements of F1 scores for each question and answer types for pairs of datasets in the first iteration in self-training stage.
The numbers and content of types vary because the categories were different in their original papers.

\begin{table}[tbp]
    \centering
    \resizebox{\linewidth}{!}{
    \begin{tabular}{llccccc}
        \toprule
        \multicolumn{2}{c}{\textbf{Dataset}} & \multicolumn{5}{c}{\textbf{Threshold $\theta$}} \\
        \cmidrule(lr){1-2} \cmidrule(lr){3-7}
        Source & Target & \multirow{1}{*}{0.5} & \multirow{1}{*}{0.6} & \multirow{1}{*}{0.7} & \multirow{1}{*}{0.8} & \multirow{1}{*}{0.9} \\
        \midrule
        \multirow{3}{*}{SQuAD} & FQuAD & 75.21 & 75.66 & 75.75 & 74.07 & 73.01\\
        {} & DRCD & 86.07 & 86.10 & 86.64 & 85.80 & 85.62 \\
        {} & KorQuAD & 83.19 & 83.46 & 83.99 & 83.79 & 82.80 \\
        \bottomrule
    \end{tabular}
    }
    \caption{
        Comparison of the F1 scores on different datasets when different thresholds were used in pseudo-labeling.
    }
    \label{tab:diff-threshold}
\end{table}

\section{Confidence Threshold}
Table~\ref{tab:diff-threshold} lists the F1 scores of the models which were fine-tuned on SQuAD then self-trained on other datasets with different thresholds in pseudo-labeling.
The best performances were obtained when $\theta = 0.7$.

\end{document}